\documentclass[sigconf,nonacm]{acmart}

\usepackage{amsmath,amsfonts,bm,relsize,dsfont,color}
\usepackage{multirow}

\usepackage{multirow}
\definecolor{MyDarkBlue}{rgb}{0,0.08,1}
\definecolor{MyDarkGreen}{rgb}{0.02,0.6,0.02}
\definecolor{MyDarkRed}{rgb}{0.8,0.02,0.02}
\definecolor{MyDarkOrange}{rgb}{0.40,0.2,0.02}
\definecolor{MyPurple}{RGB}{111,0,255}
\definecolor{MyRed}{rgb}{1.0,0.0,0.0}
\definecolor{MyGold}{rgb}{0.75,0.6,0.12}
\definecolor{MyDarkgray}{rgb}{0.66, 0.66, 0.66}
\definecolor{MyDarkCyan}{rgb}{0.05, 0.55, 0.45}
\definecolor{MyBlack}{rgb}{0., 0., 0.}
\definecolor{MyMagenta}{rgb}{1., 0., 1.}
\definecolor{BerkeleyYellow}{RGB}{255,204,41}
\definecolor{BerkeleyLightBlue}{RGB}{94,146,221}
\definecolor{BkDarkBlue}{rgb}{.05,.07,.353}

\newcommand{\MYhref}[3][blue]{\href{#2}{\color{#1}{#3}}} 

\def\ie{\textit{i.e.}}
\def\eg{\textit{e.g.}}
\def\etc{\textit{etc}}
\def\etal{\textit{et al.}}

\definecolor{NewGreen}{rgb}{.43,.67,.27}
\definecolor{NewOrange}{rgb}{.92,.49,.19}
\definecolor{NewBlue}{rgb}{.26,.44,.76}

\newcommand{\RR}{\mathbb{R}}
\newcommand{\codebook}{\mathcal{Z}}
\newcommand{\discriminator}{D}
\newcommand{\decoder}{G}
\newcommand{\encoder}{E}
\newcommand{\quantize}{\mathbf{q}}
\newcommand{\quantizedcode}{z_{\mathbf{q}}}
\newcommand{\codebookdim}{n_z}
\DeclareMathOperator*{\argmin}{arg\,min}

\newcommand{\decodersal}{G_{\text{Sal}}}

\newlength\savewidth\newcommand\shline{\noalign{\global\savewidth\arrayrulewidth
  \global\arrayrulewidth 1pt}\hline\noalign{\global\arrayrulewidth\savewidth}}
\newcommand{\tablestyle}[2]{\setlength{\tabcolsep}{#1}\renewcommand{\arraystretch}{#2}\centering\footnotesize}
\newcommand{\FigI}{
\begin{figure}
    \centering
    \includegraphics[width=1\linewidth]{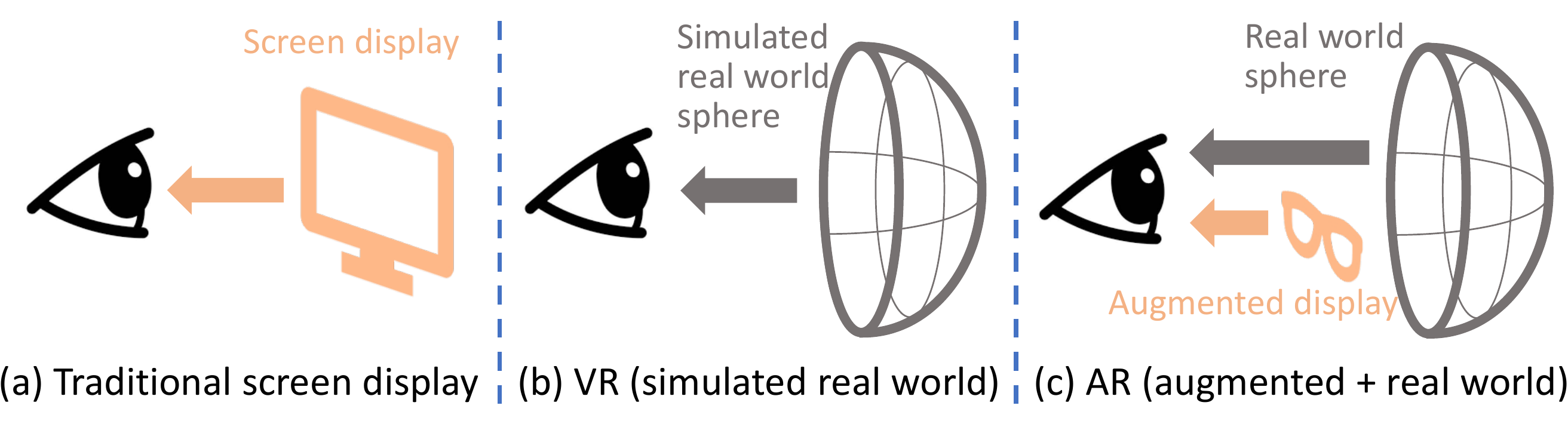}
    \vspace{-20pt}
    \caption{Illustration of the perception differences between traditional screen display, VR display, and AR display.}
    \vspace{-10pt}
    \label{fig:1}
\end{figure}
}

\newcommand{\FigII}{
\begin{figure*}
    \centering
    \includegraphics[width=1\linewidth]{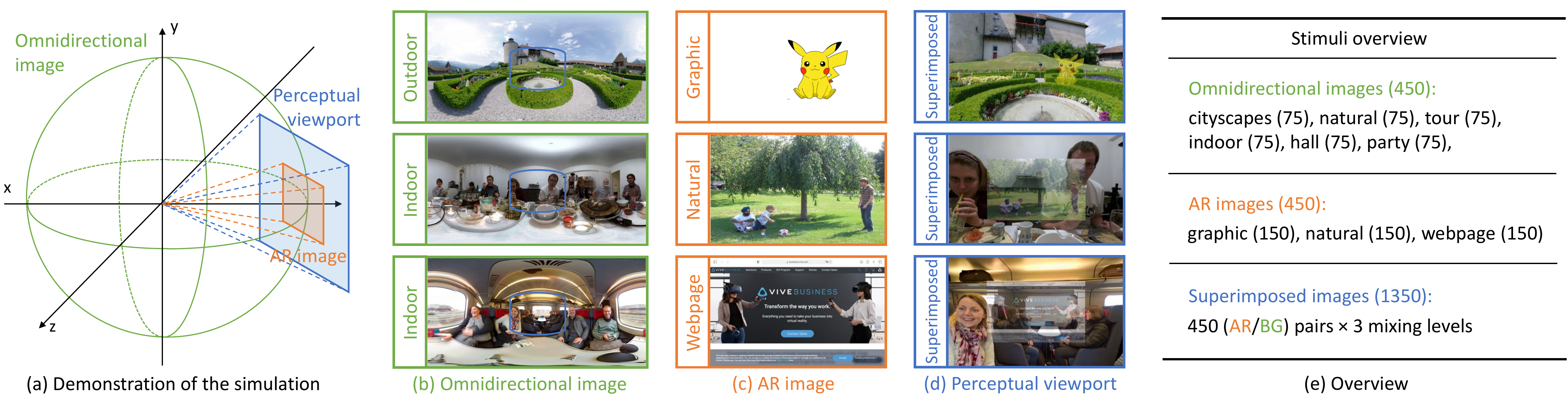}
    \vspace{-18pt}
    \caption{The illustration of the AR simulation in VR environment. (a) The demonstration of the relationship between the \textcolor{NewGreen}{omnidirectional image}, the \textcolor{NewOrange}{AR image}, and the \textcolor{NewBlue}{perceptual viewport image}. (b) Examples of the \textcolor{NewGreen}{omnidirectional images}. (c) Examples of the \textcolor{NewOrange}{AR images}. (d) Examples of the \textcolor{NewBlue}{perceptual viewport images}. Note that the perceptual viewports of the subjects are changed dynamically with the head movement, however, the relative positional relationship between the omnidirectional image and the AR image is fixed. (e) An overview of the stimuli in our dataset.}
    \vspace{-5pt}
    \label{fig:2}
\end{figure*}
}

\newcommand{\FigIII}{
\begin{figure*}
    \centering
    \includegraphics[width=0.98\linewidth]{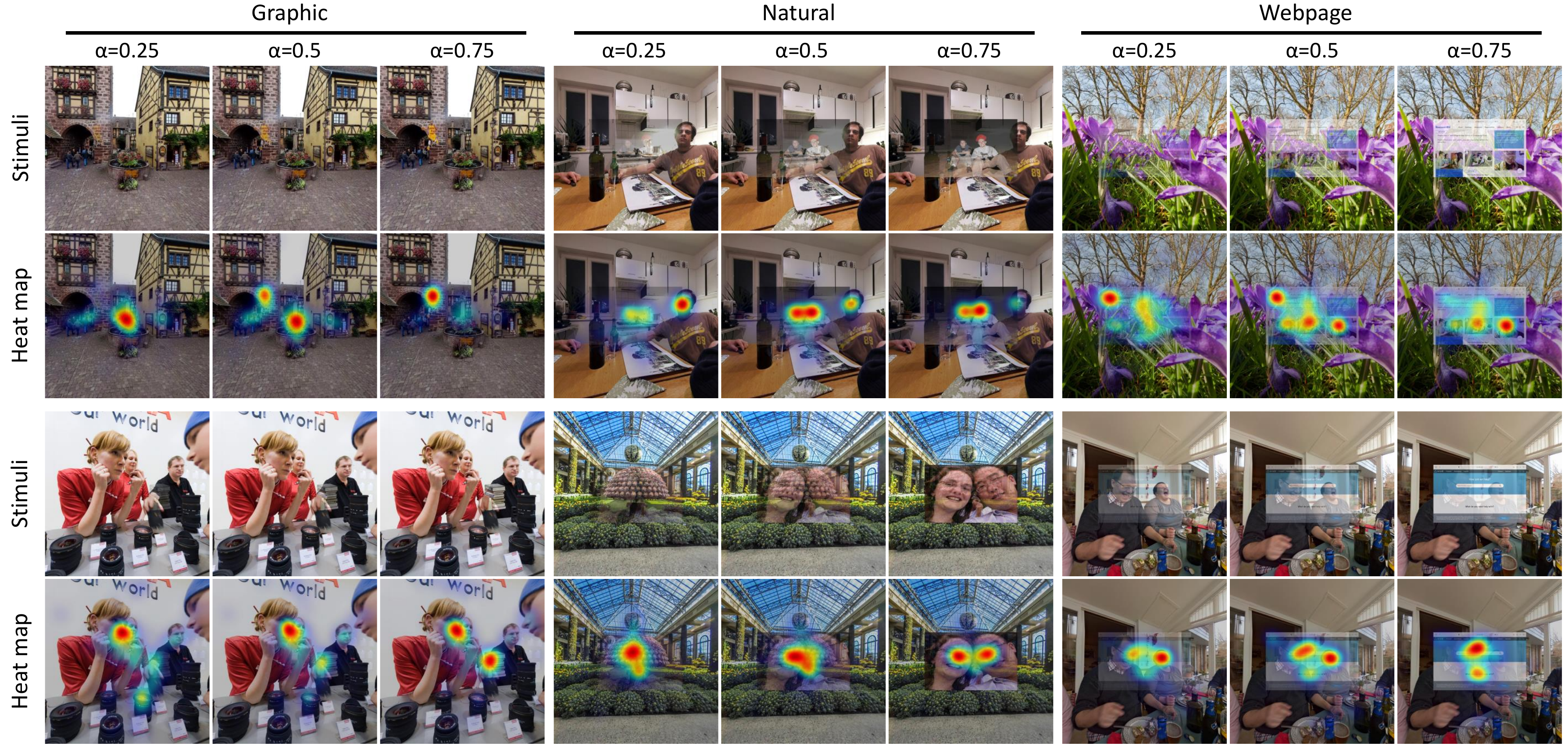}
    \vspace{-5.6pt}
    \caption{Qualitative comparisons of saliency maps for stimuli with various mixing values. The augmented contents can be derived from the comparison between stimuli with different mixing values.}
    \vspace{-5pt}
    \label{fig:3}
\end{figure*}
}

\newcommand{\FigIV}{
\begin{figure}
    \centering
    \includegraphics[width=1\linewidth]{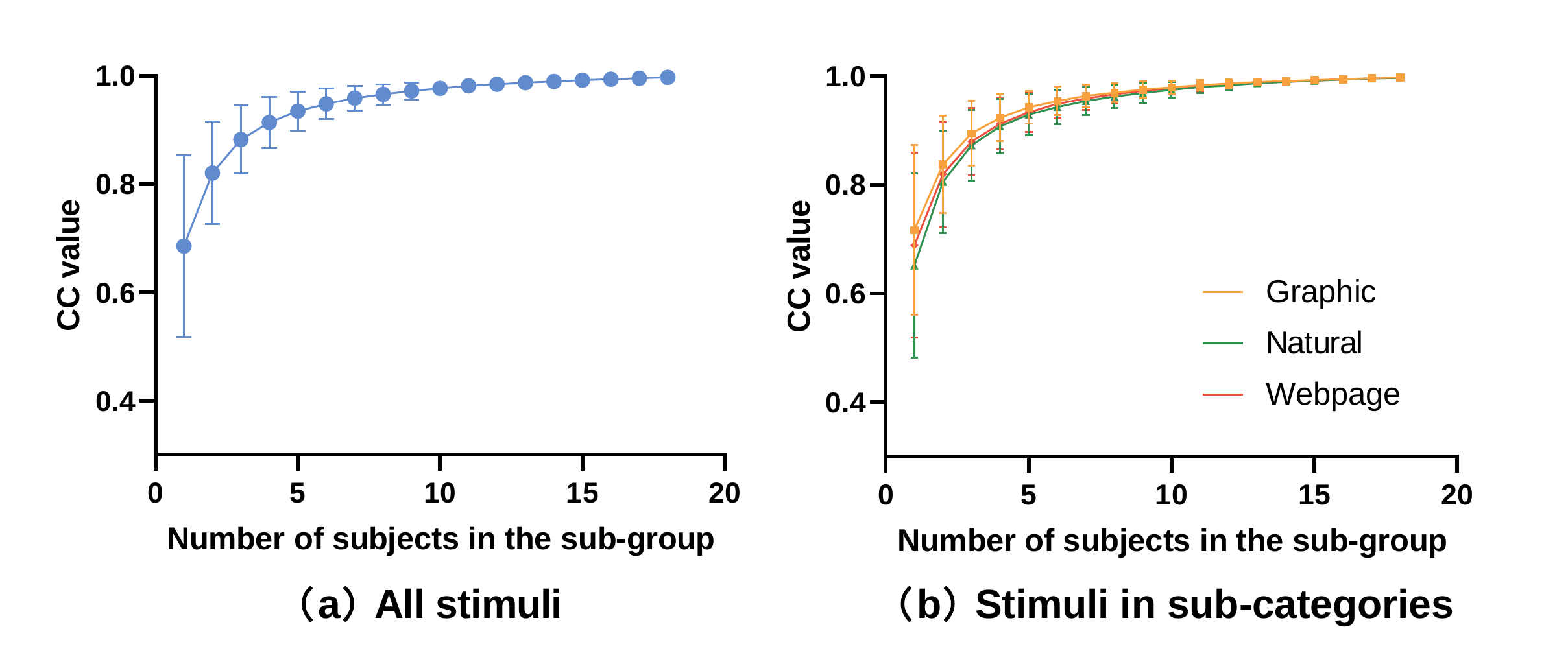}
    \vspace{-5pt}
    \caption{CC values alongside increased numbers of subjects per scenario over our SAR dataset. (a) CC values between sub-group saliency maps and overall saliency maps for all stimuli. (b) CC values between sub-group saliency maps and overall saliency maps for stimuli in sub-categories.}
    \vspace{-5pt}
    \label{fig:4}
\end{figure}
}

\newcommand{\FigV}{
\begin{figure}
    \centering
    \includegraphics[width=1\linewidth]{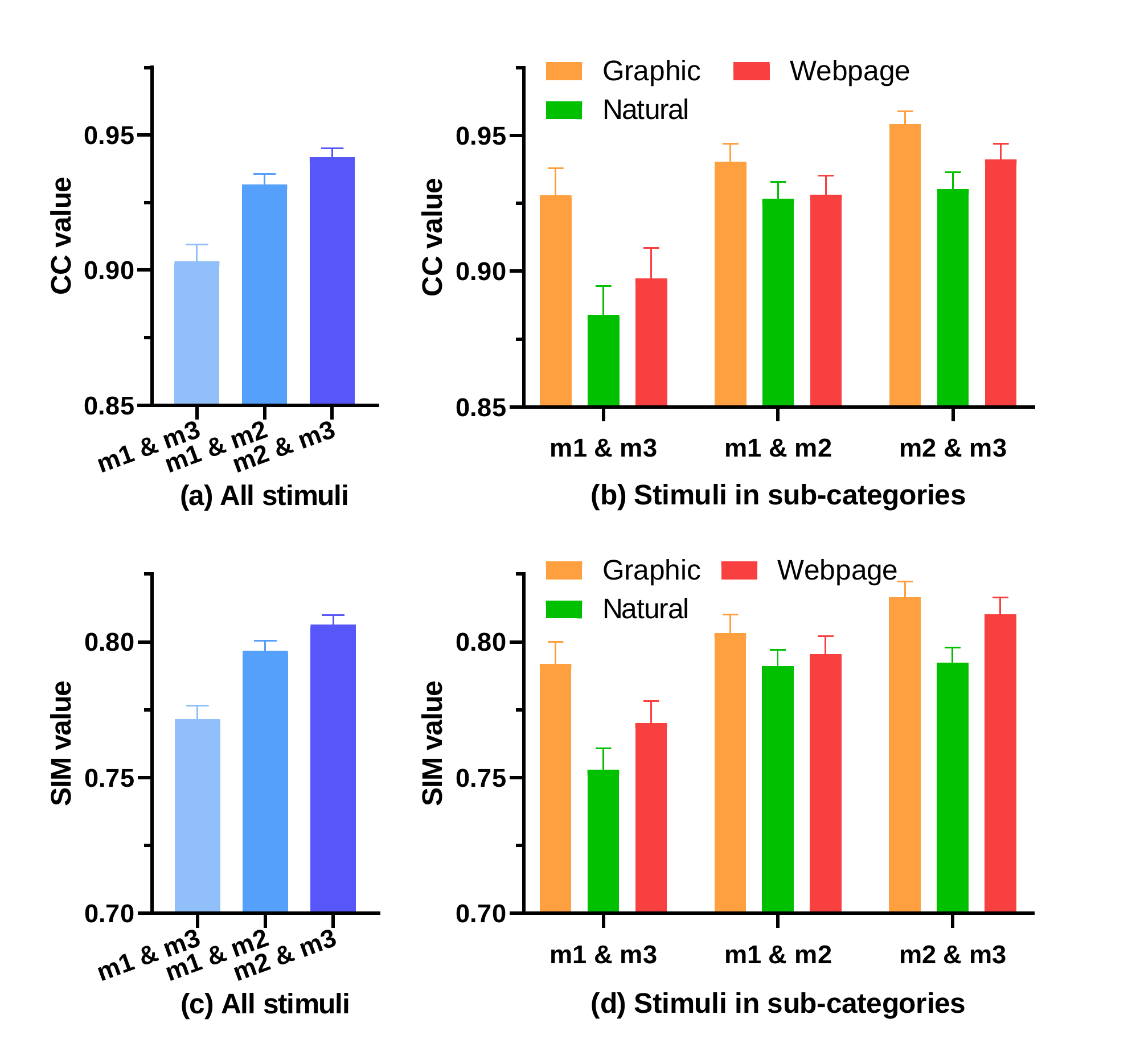}
    \vspace{-10pt}
    \caption{CC and SIM comparisons between different mixing levels (m1: $\alpha=0.25$; m2: $\alpha=0.5$; m3: $\alpha=0.75$). (a) CC comparisons for all stimuli. (b) CC comparisons for stimuli in sub-categories of augmented contents. (c) SIM comparisons for all stimuli. (d) SIM comparisons for stimuli in sub-categories of augmented contents.}
    \label{fig:5}
\end{figure}
}

\newcommand{\FigVI}{
\begin{figure}
    \centering
    \includegraphics[width=1\linewidth]{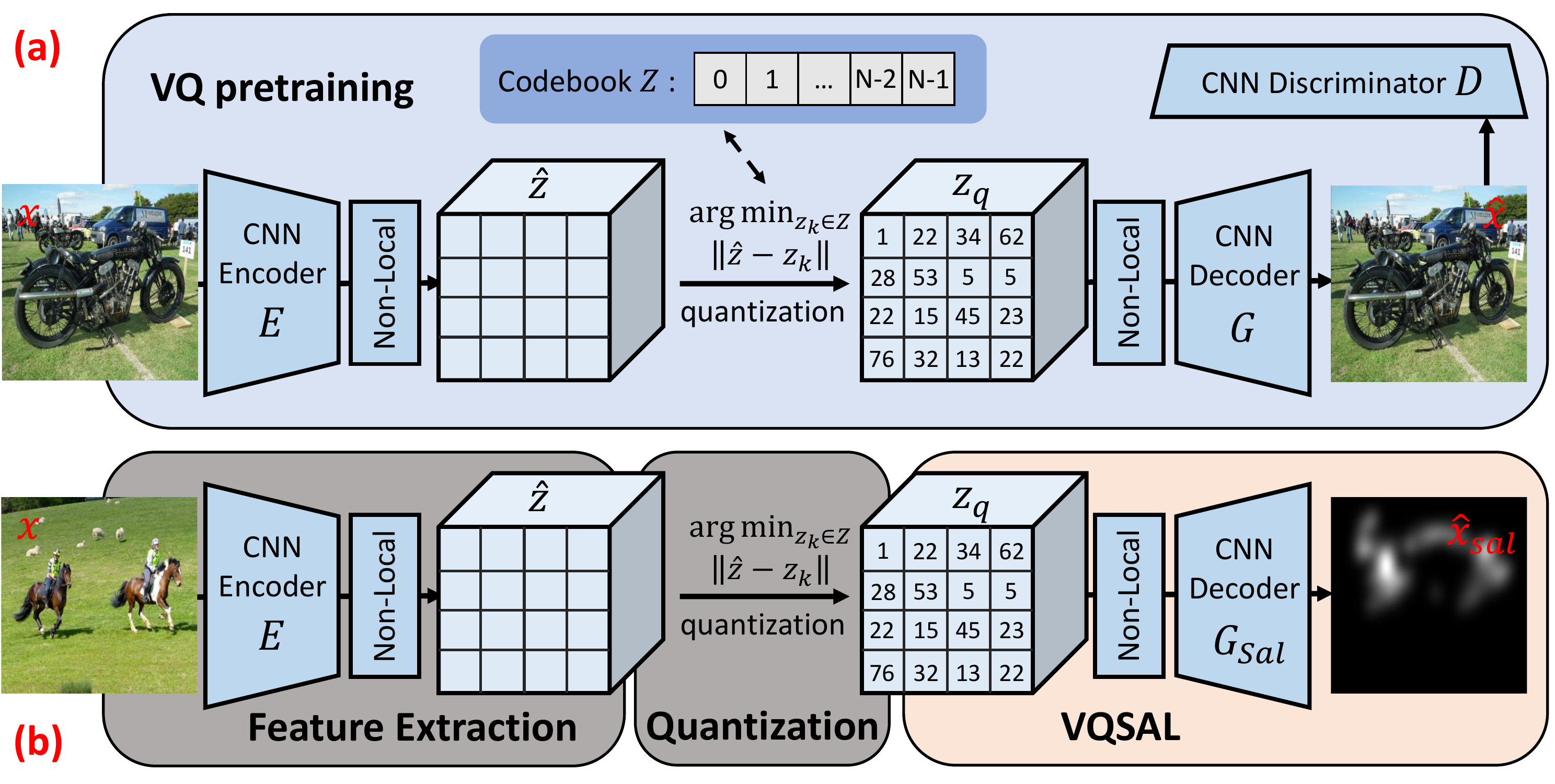}
    \vspace{-18pt}
    \caption{Overview of the proposed VQSal model. Our VQSal model is first pretrained on a large-scale unlabeled image dataset to learn a context-rich codebook for images. Then we freeze the feature extraction and quantization parts, and only finetune the decoder part to perform saliency prediction.}
    \vspace{-8pt}
    \label{fig:6}
\end{figure}
}

\newcommand{\FigVII}{
\begin{figure}
    \centering
    \includegraphics[width=1\linewidth]{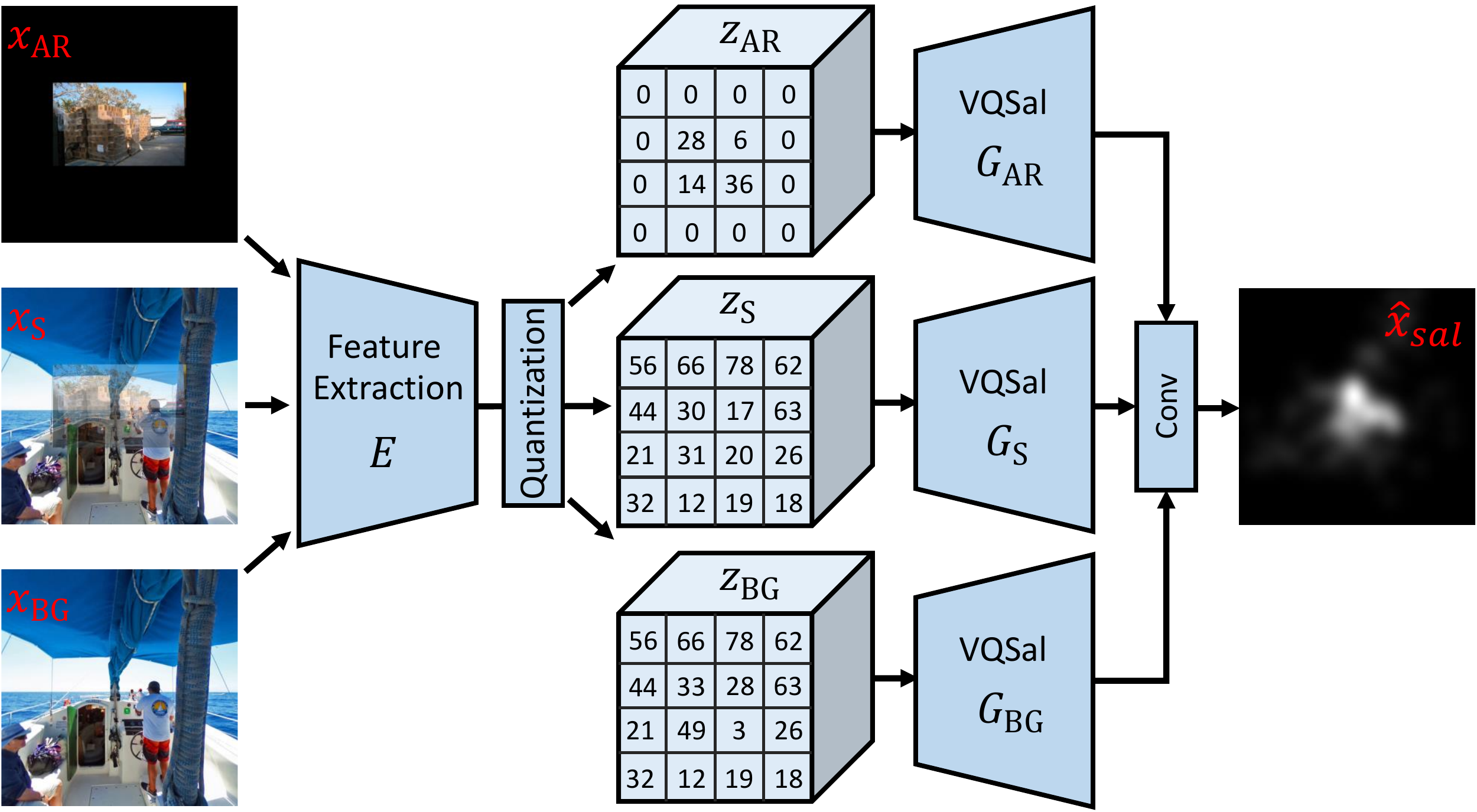}
    \vspace{-14pt}
    \caption{Overview of the proposed VQSal-AR model. Similar to the VQSal model, we first extract the features for three images (AR image, BG image, and superimposed image), and then quantize them to get the corresponding representation codes. These codes are decoded and integrated to predict AR saliency.}
    \vspace{-1pt}
    \label{fig:7}
\end{figure}
}

\newcommand{\FigVIII}{
\begin{figure*}
    \centering
    \includegraphics[width=0.98\linewidth]{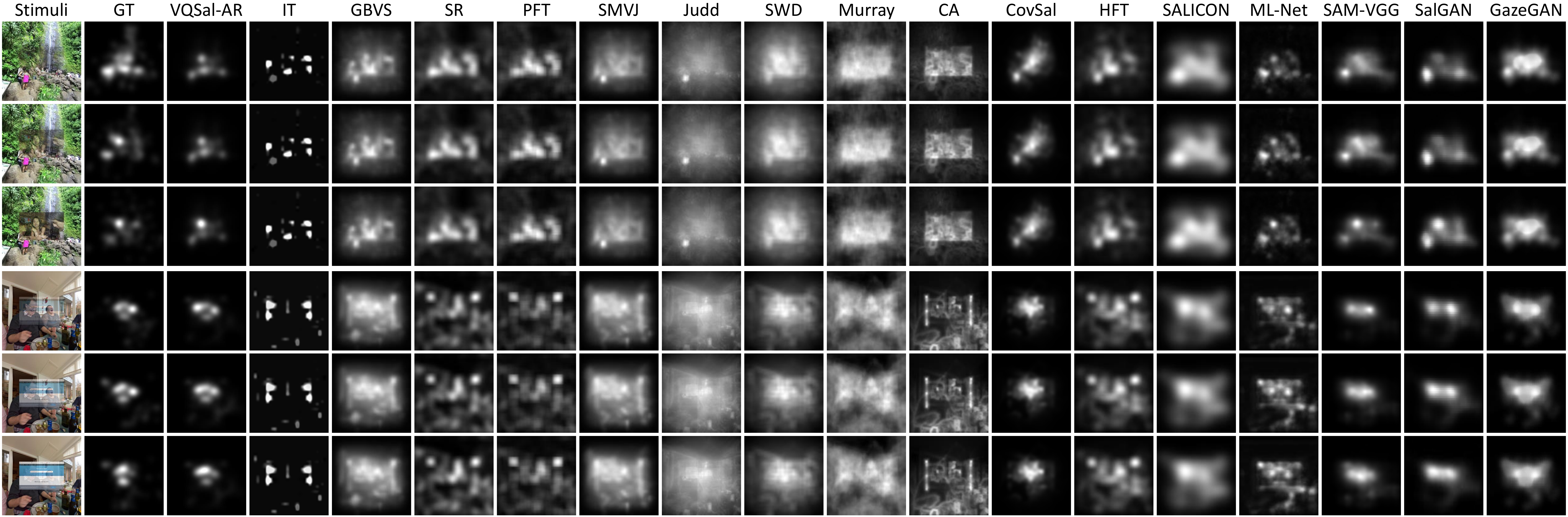}
    \vspace{-8pt}
    \caption{Qualitative comparisons for different models on our SARD.}
    \vspace{-10pt}
    \label{fig:8}
\end{figure*}
}
\newcommand{\TabI}{
\begin{table}
  \setlength{\tabcolsep}{3.4pt}
  \caption{Quantitative comparison results of different models on the Salicon \cite{jiang2015salicon} dataset. We {\fontsize{11pt}{\baselineskip}\selectfont bold} the best result for each metric.}
  \vspace{-8pt}
  \label{tab:1_salicon}
  \begin{tabular}{l|cccccc}
    \toprule
    Model $\backslash$ Metric	&	AUC $\uparrow$	&	CC $\uparrow$	&	 IG $\uparrow$	&	KL $\downarrow$	&	 NSS $\uparrow$	&	SIM $\uparrow$	\\
    \midrule 
    SALICON \cite{huang2015salicon}	&	0.824	&	0.716	&	34.77	&	5.723	&	1.404	&	0.637	\\
    ML-Net \cite{cornia2016deep}	&	0.809	&	0.675	&	34.79	&	5.707	&	1.509	&	0.591	\\
    SAM-VGG \cite{cornia2018predicting}	&	0.856	&	0.836	&	34.98	&	5.571	&	1.802	&	0.738	\\
    SAM-ResNet \cite{cornia2018predicting}	&	0.855	&	0.844	&	35.04	&	5.535	&	1.809	&	0.738	\\
    SalGAN \cite{pan2017salgan}	&	0.858	&	0.848	&	35.16	&	5.449	&	1.763	&	0.743	\\
    GazeGAN \cite{che2019gaze}	&	0.853	&	0.826	&	34.71	&	5.760	&	1.736	&	0.732	\\
    VQSal (Ours)	&	\textbf{0.863}	&	\textbf{0.869}	&	\textbf{35.18}	&	\textbf{5.434}	&	\textbf{1.863}	&	\textbf{0.766}	\\
    \bottomrule
  \end{tabular}
  \vspace{-5pt}
\end{table}
}

\newcommand{\TabII}{
\begin{table*}
  \renewcommand{\arraystretch}{0.988}
  \setlength{\tabcolsep}{3.1pt}
  \caption{Quantitative comparisons between our proposed VQSal-AR model and three types of benchmark methods. We {\fontsize{11pt}{\baselineskip}\selectfont bold} the best result and \underline{underline} the second-best result for each metric in each type. The best result for each metric throughout the table is colored in \textcolor{red}{red} and the second-best result for each metric throughout the table is colored in \textcolor{blue}{blue}.}
  \vspace{-5pt}
  \label{tab:2_ar}
  \begin{tabular}{l|ccccc|ccccc|ccccc}
    \toprule
    Type	&	\multicolumn{5}{c}{Type I}	\vline &	\multicolumn{5}{c}{Type II}	\vline &	\multicolumn{5}{c}{Type III}\\
    
    \midrule
    Model $\backslash$ Metric	&	AUC $\uparrow$	&	sAUC $\uparrow$	&	CC $\uparrow$	&	 NSS $\uparrow$	&	SIM $\uparrow$	&	AUC $\uparrow$	&	sAUC $\uparrow$	&	CC $\uparrow$	&	 NSS $\uparrow$	&	SIM $\uparrow$	&	AUC $\uparrow$	&	sAUC $\uparrow$	&	CC $\uparrow$	&	 NSS $\uparrow$	&	SIM $\uparrow$	\\
    
    \midrule 
    IT \cite{itti1998model}	&	0.541	&	0.509	&	0.075	&	0.272	&	0.105	&	0.628	&	0.528	&	0.262	&	0.859	&	0.238	&	0.622	&	0.533	&	0.308	&	1.004	&	0.369	\\
    AIM \cite{bruce2005saliency}	&	0.701	&	\underline{0.557}	&	0.237	&	0.689	&	0.371	&	0.826	&	0.559	&	0.534	&	1.441	&	0.468	&	0.838	&	0.551	&	0.572	&	1.531	&	0.467	\\
    GBVS \cite{harel2006graph}	&	0.766	&	0.510	&	0.332	&	0.848	&	0.424	&	0.850	&	0.547	&	0.601	&	1.623	&	0.515	&	0.853	&	0.550	&	0.595	&	1.575	&	0.510	\\
    SR \cite{hou2007saliency}	&	0.637	&	0.513	&	0.120	&	0.359	&	0.329	&	0.791	&	0.550	&	0.458	&	1.279	&	0.456	&	0.811	&	0.552	&	0.522	&	1.443	&	0.470	\\
    SUN \cite{zhang2008sun}	&	0.679	&	0.534	&	0.188	&	0.640	&	0.352	&	0.806	&	\textbf{0.568}	&	0.491	&	1.462	&	0.445	&	0.827	&	\textbf{0.569}	&	0.561	&	1.635	&	0.465	\\
    PFT \cite{guo2008spatio}	&	0.634	&	0.512	&	0.107	&	0.324	&	0.327	&	0.796	&	0.551	&	0.478	&	1.342	&	0.457	&	0.820	&	0.551	&	0.569	&	1.587	&	0.484	\\
    SMVJ \cite{cerf2007predicting}	&	0.803	&	0.529	&	0.439	&	1.123	&	0.457	&	0.853	&	0.549	&	0.611	&	1.626	&	0.520	&	0.853	&	0.549	&	0.584	&	1.523	&	0.503	\\
    Judd \cite{judd2009learning}	&	\underline{0.863}	&	0.531	&	0.541	&	1.415	&	0.416	&	\underline{0.878}	&	0.536	&	0.637	&	1.680	&	0.447	&	\textbf{0.881}	&	0.529	&	0.601	&	1.563	&	0.430	\\
    SWD \cite{duan2011visual}	&	0.853	&	0.540	&	\underline{0.566}	&	\underline{1.483}	&	\underline{0.466}	&	\textbf{0.880}	&	0.554	&	\underline{0.722}	&	\underline{1.946}	&	0.528	&	\underline{0.880}	&	0.545	&	\underline{0.693}	&	\underline{1.830}	&	0.507	\\
    Murray \cite{murray2011saliency}	&	0.650	&	\textbf{0.559}	&	0.181	&	0.520	&	0.356	&	0.829	&	0.549	&	0.521	&	1.364	&	0.443	&	0.847	&	0.544	&	0.573	&	1.485	&	0.454	\\
    CA \cite{goferman2011context}	&	0.707	&	0.533	&	0.231	&	0.674	&	0.378	&	0.822	&	\underline{0.564}	&	0.528	&	1.488	&	0.486	&	0.832	&	\underline{0.563}	&	0.558	&	1.558	&	0.488	\\
    CovSal \cite{erdem2013visual}	&	\textbf{0.864}	&	0.520	&	\textbf{0.695}	&	\textbf{1.859}	&	\textbf{0.608}	&	0.872	&	0.538	&	\textbf{0.768}	&	\textbf{2.143}	&	\textbf{0.649}	&	0.872	&	0.532	&	\textbf{0.751}	&	\textbf{2.035}	&	\textbf{0.641}	\\
    HFT \cite{li2012visual}	&	0.765	&	0.522	&	0.347	&	0.937	&	0.429	&	0.857	&	0.554	&	0.646	&	1.788	&	\underline{0.551}	&	0.860	&	0.559	&	0.654	&	1.795	&	\underline{0.548}	\\
    
    \midrule
    SALICON \cite{huang2015salicon}	&	0.828	&	0.567	&	0.520	&	1.364	&	0.499	&	0.867	&	0.563	&	0.655	&	1.725	&	0.557	&	0.871	&	0.567	&	0.651	&	1.706	&	0.549	\\
    ML-Net \cite{cornia2016deep}	&	0.853	&	\textbf{0.579}	&	0.655	&	2.141	&	0.569	&	0.856	&	\underline{0.586}	&	0.642	&	1.943	&	0.506	&	0.877	&	0.590	&	0.728	&	2.259	&	0.561	\\
    SAM-VGG \cite{cornia2018predicting}	&	\underline{0.892}	&	0.555	&	\underline{0.846}	&	\underline{2.412}	&	\underline{0.716}	&	0.881	&	0.580	&	\underline{0.747}	&	\underline{2.102}	&	0.635	&	0.897	&	0.577	&	\underline{0.840}	&	2.362	&	0.690	\\
			
    SalGAN \cite{pan2017salgan}	&	0.887	&	0.573	&	0.765	&	2.111	&	0.662	&	0.881	&	0.580	&	\textbf{0.751}	&	2.068	&	\underline{0.636}	&	0.891	&	0.585	&	0.782	&	2.150	&	0.658	\\
    GazeGAN \cite{che2019gaze}	&	0.887	&	0.548	&	0.809	&	2.219	&	0.661	&	\underline{0.882}	&	0.576	&	0.741	&	2.058	&	0.631	&	0.894	&	0.571	&	0.804	&	2.214	&	0.669	\\
    VQSal (Ours)	&	\textbf{0.899}	&	\underline{0.576}	&	\textbf{\textcolor{blue}{0.873}}	&	\textbf{\textcolor{blue}{2.606}}	&	\textbf{\textcolor{blue}{0.741}}	&	\textbf{0.885}	&	\textbf{0.587}	&	\underline{0.747}	&	\textbf{2.120}	&	\textbf{0.644}	&	\underline{\textcolor{blue}{0.900}}	&	\textbf{\textcolor{red}{0.597}}	&	\underline{0.840}	&	\underline{2.419}	&	\underline{0.697}	\\
    VQSal-AR (Ours)	&	-	&	-	&	-	&	-	&	-	&	-	&	-	&	-	&	-	&	-	&	\textbf{\textcolor{red}{0.903}}	&	\underline{\textcolor{blue}{0.591}}	&	\textbf{\textcolor{red}{0.893}}	&	\textbf{\textcolor{red}{2.687}}	&	\textbf{\textcolor{red}{0.758}}	\\

    \bottomrule
  \end{tabular}
  \vspace{-8pt}
\end{table*}
}

\begin{document}

\title{Saliency in Augmented Reality}

\author{Huiyu Duan$^1$, Wei Shen$^1$, Xiongkuo Min$^1$, Danyang Tu$^1$, Jing Li$^2$ and Guangtao Zhai$^1$}
\email{{huiyuduan, wei.shen, minxiongkuo, danyangtu, zhaiguangtao}@sjtu.edu.cn,  lj225205@alibaba-inc.com}
\affiliation{%
  \institution{$^1$ Shanghai Jiao Tong University, Shanghai, China \\
  $^2$ Alibaba Group, Hangzhou, China}
  \country{}
}

\renewcommand{\shortauthors}{Huiyu Duan et al.}

\begin{abstract}

With the rapid development of multimedia technology, Augmented Reality (AR) has become a promising next-generation mobile platform.
The primary theory underlying AR is human visual confusion, which allows users to perceive the real-world scenes and augmented contents (virtual-world scenes) simultaneously by superimposing them together.
To achieve good Quality of Experience (QoE), it is important to understand the interaction between two scenarios, and harmoniously display AR contents.
However, studies on how this superimposition will influence the human visual attention are lacking.
Therefore, in this paper, we mainly analyze the interaction effect between background (BG) scenes and AR contents, and study the saliency prediction problem in AR.
Specifically, we first construct a Saliency in AR Dataset (SARD), which contains 450 BG images, 450 AR images, as well as 1350 superimposed images generated by superimposing BG and AR images in pair with three mixing levels.
A large-scale eye-tracking experiment among 60 subjects is conducted to collect eye movement data.
To better predict the saliency in AR, we propose a vector quantized saliency prediction method and generalize it for AR saliency prediction.
For comparison, three benchmark methods are proposed and evaluated together with our proposed method on our SARD.
Experimental results demonstrate the superiority of our proposed method on both of the common saliency prediction problem and the AR saliency prediction problem over benchmark methods.
Our dataset and code are available at: \MYhref[MyMagenta]{https://github.com/DuanHuiyu/ARSaliency}{https://github.com/DuanHuiyu/ARSaliency}.

\end{abstract}


\maketitle

\section{Introduction \label{sec:1_introduction}}


With the evolution of multimedia technology, head-mounted display (HMD) technologies (\eg, Virtual Reality (VR), Augmented Reality (AR), Mixed Reality (MR), \etc.) have attracted more attention recently and have become promising next-generation display solutions \cite{cakmakci2006head,zhan2020augmented}.
Among these technologies, AR aims at enriching the real-world information by superimposing virtual contents on it, which promises to be the next-generation mobile platform.
Since AR technology can display augmented contents while keeping real-world information, it has great potentials in many applications, such as communication, entertainment, health care, education, engineering design, \etc.

\FigI

The superimposition of background (BG) scenes (\ie, real-world scenes) and AR contents (\ie, virtual-world contents) in AR display will cause visual confusion \cite{woods2010extended,peli2017multiplexing}, which alters the way that humans perceive both of the real world and virtual world \cite{duan2022confusing,duan2022unified,duan2022develop}.
The primary value of AR is to enrich the real-world information, however, inappropriate augmented design may affect the Quality of Experience (QoE) of users \cite{duan2022confusing}, arouse inattentional blindness \cite{mack2003inattentional}, and may even cause danger \cite{wang2021inattentional}.
Therefore, it is important to analyze and model the visual attention behaviour of humans in AR environment.

Visual attention analysis and prediction have long been important tasks in both multimedia and computer vision research \cite{itti1998model,rai2017dataset,borji2019saliency}, since they can give new insights about human attention mechanisms and contribute to many multimedia applications, such as image compression, object and motion detection, \etc.
Many previous works on saliency datasets \cite{bylinskii2015saliency,jiang2015salicon} and saliency prediction models \cite{judd2009learning,che2019gaze} have been conducted on screen display as shown in Figure \ref{fig:1} (a).
Recently, with the popularity of head-mounted displays, some visual attention datasets \cite{rai2017dataset,david2018dataset} and saliency prediction models \cite{zhang2021graph} towards VR technology have been proposed as demonstrated in Figure \ref{fig:1} (b). 
Most of these works focused on the saliency tasks on omnidirectional images/videos \cite{rai2017dataset}, or egocentric videos \cite{zhang2017deep}.
Although AR technology has great potentials, most of existing AR studies mainly focus on egocentric video understanding \cite{grauman2021ego4d}, while they may ignore that the augmented contents are also important for AR.
Zhu \etal~\cite{zhu2019saliency} have constructed a saliency dataset for omnidirectional videos with augmented bounding box contents on them.
However, as shown in Figure \ref{fig:2} (a), it is hard to acquire real-world omnidirectional view in AR applications, and augmented contents are usually complex and have various superimposition degrees rather than overlaying simple bounding boxes.
Duan \etal~\cite{duan2022confusing} have studied the visual confusion effect in AR technology.
However, they only explored the influence of visual confusion on quality of experience (QoE), while the visual attention research is still lacking.
As shown in Figure \ref{fig:1} (c), the visual confusion caused by the superimposition of AR contents and BG scenes is significant in AR, however, the understanding of saliency in Augmented Reality environment is still limited.

Modeling the visual attention in AR can help better design, display, and adaptively adjust the virtual contents to accord the expectation of human vision \cite{kruijff2010perceptual}, as well as contribute to AR QoE assessment \cite{duan2022confusing} and augmented image compression method design \cite{zhu2019saliency} \etc.
Therefore, in this work, we aim at analyzing the human visual attention behavior in AR thoroughly and building an accurate saliency prediction paradigm for AR.
To achieve this objective, we are facing the following research challenges:

\textbf{(i)} \textit{Building a dataset for AR saliency.}
Although there are many famous AR products, \eg, HoloLens \cite{hololens}, Magic Leap \cite{magicleap}, Epson AR \cite{epson}, \etc., it is hard to conduct controllable eye-tracking experiments with these devices in real scenes.
Moreover, the experimental scenarios may also be limited by the experimental environment.

\textbf{(ii)} \textit{Understanding the effect of visual confusion on AR saliency.}
As a common observation, a higher opacity value for augmented contents may make them clearer and attract more attention, while a lower opacity value for augmented contents may make background scenes clearer and attract more attention.
Despite this general consensus, we observe that our understanding of the influence of the mixing level is still limited, and this general consensus cannot be generalized to various complex applications.

\textbf{(iii)} \textit{Modeling AR saliency.}
Since humans can perceive two layers of images in Augmented Reality, \ie, an AR image and a BG image, it is necessary and significant to study how to jointly exploit these two parts of information to build an accurate saliency model for AR.

In this work, to solve the dataset challenge, we first propose to conduct the AR eye-tracking experiment under the VR environment.
As shown in Figure \ref{fig:1} (b) and (c), since VR is used to simulate real world scenes, it is also competent to be used as BG scenes in the AR experiments.
Specifically, we collect 450 omnidirectional images as BG images, and 450 common images including 150 graphic images, 150 natural images, 150 webpage images as AR images.
To better understand the influence of mixing levels on the saliency in AR, three mixing levels are used during the eye tracking experiment, and the experiment of each mixing level is conducted among different subjects.
To better predict the visual attention in AR, we propose a vector quantization-based saliency model and utilize multi-decoders to integrate obtained information for joint prediction.
For comparison, three benchmark paradigms for AR saliency prediction are proposed and evaluated based on our dataset.
Experimental results demonstrate that our proposed model achieves state-of-the-art performance compared to other baseline methods.
The contributions are summarized as follows.

\begin{itemize}
    \item We build the first AR saliency dataset that considers the visual confusion effect in AR.
    \item We analyze the influence of visual confusion on visual attention for various stimuli. 
    \item A saliency prediction model for predicting AR saliency is proposed.
    \item Three benchmark methods are proposed and evaluated based on our dataset, and our proposed model achieves state-of-the-art performance.
\end{itemize}

\section{Related Work}

\subsection{Eye-tracking Datasets}

\hspace{0.91em} \textit{\textbf{Traditional Saliency Datasets.}} Humans have remarkable abilities to search and focus on salient regions in a scene \cite{marois2005capacity,bundesen2015recent}, which allows us to efficiently process a large amount of information.
This neural mechanism is known as visual attention.
To understand and model visual attention behavior, many eye-tracking datasets have been constructed.
MIT1003 \cite{judd2009learning} is a large-scale saliency dataset, which contains 1003 images.
MIT300 and CAT2000 \cite{bylinskii2015saliency} are two widely used benchmark datasets, which contains 300 and 2000 test images respectively.
SALICON \cite{jiang2015salicon} is currently the largest crowd-sourced saliency dataset, which contains 10000 training images, 5000 validation images and 5000 test images collected through mouse tracking using Amazon Mechanical Turk (AMT).
This dataset is widely used to pretrain saliency models.

\textit{\textbf{VR/AR Saliency Datasets.}}
Recently, with the popularity of HMDs and the concept of XR, many eye-tracking datasets have been constructed towards the applications in these new display technologies \cite{zhu2021viewing}.
Salient360 \cite{rai2017dataset} is one of the earliest omnidirectional datasets for saliency prediction, which contains 98 stimuli including indoor, outdoor and people scenes.
For each omnidirectional image, at least 40 subjects were recruited to view the stimuli for 25 seconds.
AOI \cite{xu2021saliency} is a large-scale omnidirectional saliency dataset, which contains 600 images and the corresponding head/eye fixations obtained from 30 subjects.
All of these datasets are constructed towards the VR saliency detection task.
Zhu \etal~\cite{zhu2019saliency} have established a saliency dataset for 50 omnidirectional videos with bounding boxes.
However, this is still an omnidirectional saliency task.
Moreover, the bounding box contents can only cover part of AR applications, and they only overlay the AR contents on BG scenes, which ignores the mixing value between AR and BG.
In this work, we argue that the interaction between AR contents and BG scenes is important in AR display, and study this important task among various scenarios within perceptual viewport images.

\vspace{-3pt}
\subsection{Saliency Prediction Models}

\hspace{0.91em} \textit{\textbf{Classical Saliency Models.}} 
Most traditional methods have modeled visual saliency based on the bottom-up mechanism.
The early models mainly relied on extracting simple low-level feature maps such as intensity, color, and orientation \etc~\cite{itti1998model}.
Some subsequent models incorporated middle-level and high-level features to better predict visual attention \cite{liang2015predicting,judd2009learning}.
These classical methods including Attention for Information Maximization (AIM) \cite{bruce2005saliency}, Graph-based Visual Saliency (GBVS) \cite{harel2006graph}, Judd model \cite{judd2009learning}, \etc., are still highly influential in current visual attention research.

\FigII

\textit{\textbf{Deep Saliency Models.}} 
With the development of deep neural network (DNN), the saliency prediction task has achieved significant improvement recently \cite{vig2014large,wang2015deep,kruthiventi2017deepfix,wang2016saliency,tu2022end}.
Huang \etal~\cite{huang2015salicon} proposed a two stream convolutional neural network (CNN) to extract coarse and fine features to compute saliency map.
Cornia \etal~\cite{cornia2018predicting} used long short-term memory (LSTM) to enhance the extracted feature maps from a dilated CNN to predict saliency.
Pan \etal~\cite{pan2017salgan} proposed to use the generative adversarial network (GAN) to calculate the saliency map.
Che \etal~\cite{che2019gaze} studied the influence of transformation on visual attention and proposed a GazeGAN model based on U-Net for saliency prediction.
These models based on the top-down mechanism have been widely used in various research fields recently \cite{borji2019saliency}.

\vspace{-3pt}
\subsection{Augmented Reality}
This work mainly concerns head-mounted AR application rather than mobile phone based AR application.
Early head-mounted AR devices, \eg, Google Glass \cite{googleglass}, generally display augmented contents for only one eye and used another eye for perceiving real-world scenes based on binocular visual confusion \cite{duan2022confusing,carmigniani2011augmented}.
However, binocular rivalry caused by binocular visual confusion may strongly affect the QoE \cite{kruijff2010perceptual,duan2022confusing}.
Recently, most AR technologies are built based on monocular visual confusion to avoid occluding when displaying augmented contents, such as Microsoft HoloLens \cite{hololens}, Magic Leap \cite{magicleap}, Epson AR \cite{epson}, \textit{etc.}, since monocular rivalry is much weaker than binocular rivalry \cite{o2009monocular}.
In this paper, we mainly consider these monocular visual confusion based technologies.
This type of device generally has external cameras and internal gyroscope sensors to register the location of augmented contents in real-world environment, thus it is easy to get the position relationship between the AR contents and BG scenes \cite{zhan2020augmented,duan2022confusing}.

\section{SARD: Saliency in AR Dataset \label{sec:3}}

\subsection{Experimental Methodology \label{sec:3.1}}
In real applications, since AR usually needs external cameras on HMDs to register and locate real-world scenes for augmented rendering, thus it is possible to capture background scenes, obtain AR contents, and acquire the relationship between AR contents and BG scenes.
An intuitive way to conduct AR eye tracking experiment is wearing AR devices in various environments and then collecting eye movement data.
However, this way suffers from uncontrollable experimental environments and limited experimental scenarios \cite{duan2022confusing}.
Another way to simulate AR scenarios is using a big screen with displaying superimposed AR/BG images on it.
However, big screens cannot create immersive experience, which may not be appropriate in this study.
As discussed in Sec. \ref{sec:1_introduction} and Figure \ref{fig:1}, VR can simulate real world scenes \cite{duan2017ivqad,duan2018perceptual}, which is capable to simulate the BG scenes in AR applications.
Therefore, as demonstrated in Figure \ref{fig:2} (a), we adopt the method of conducting AR eye-tracking experiments in VR environment for controllable experimental environments and diverse experimental scenarios.

\FigIII

\subsection{Data Collection \label{sec:3.2}}

\hspace{0.91em} \textit{\textbf{Stimuli.}}
We first collect 450 omnidirectional images from \cite{xu2021saliency} as BG scenes, which contains six categories of scenes, \ie, cityscapes, natural landscapes, human tour scenes, indoor scenes, indoor hall scenes, and human party scenes.
Example images are shown in Figure \ref{fig:2} (b).
We then collect 450 common images online as AR contents, which consist of three types of images including graphic images, natural images, and webpage images as demonstrated in Figure \ref{fig:2} (c).
For graphic images, only graphic areas are non-transparent, and for natural and webpage images, the whole images are non-transparent.
The 450 omnidirectional images and 450 AR images are randomly paired to generate 1350 various perceptual scenarios with three mixing values, and the perceptual viewport images are formulated as:
\begin{equation}
    I_S = \alpha I_{\alpha }I_\text{AR} + (1-\alpha I_{\alpha })I_\text{BG},
    \label{eq:1}
\end{equation}
where $I_{AR}$ and $I_{BG}$ are AR and BG images, respectively, $I_{\alpha }$ is the intrinsic transparency matrix for AR contents, $\alpha \in \left\{0.25,0.5,0.75 \right\}$ is the mixing value to generate superimposed images.
$I_{\alpha }$ is usually a 0-1 matrix, where 0 means transparent and 1 means non-transparent.
Since perceptual viewports are usually larger than the field-of-view (FOV) of AR contents, we pad each AR image to the perceptual viewport size with 0 values for both of the color space and the transparency space, and keep the raw color and transparency values for original contents.
The examples of generated superimposed images are shown in Figure \ref{fig:2} (d).
Note that this generation method for superimposed images is only used for the construction of the dataset.
During the eye-tracking experiment, the superimposition process is conducted in Unity3D \cite{Unity} as illustrated below in ``\textit{Procedure}''.

\textit{\textbf{Apparatus.}}
We use a HTC VIVE Pro Eye \cite{HTC} as the hardware apparatus to display omnidirectional scenes and AR stimuli, as well as to collect eye movement data.
The resolution of the displays inside HTC VIVE Pro Eye is $1440 \times 1600$ pixels per eye which covers $110^{\circ}$ FOV.
The refresh rate of the displays is 90 Hz.
Moreover, this HMD has Tobii eye-tracker inside it with the sampling frequency of 90 Hz.

\textit{\textbf{Subjects.}}
Since 3 mixing levels are imposed for each AR/BG pair, subjects may remember the scenario if they have accessed it before and it may influence the reliability of collected data.
Thus, we recruit a large number of subjects, \ie, 60 subjects (20 females and 40 males), and each subject randomly watches only one mixing level of AR/BG pairs.
As a result, each subject views totally 450 scenarios without scene repeat, and each superimposed viewport is viewed by 20 subjects.
Before participating in the test, all subjects have read and signed a consent form which explained the human study.
All participants have normal or correct-to-normal visual acuity during the experiment.

\textit{\textbf{Procedure.}}
The software system is designed using Unity3D \cite{Unity} to control the experimental procedure and record all data.
For each AR/BG pair and one mixing value $\alpha$, we set the omnidirectional (BG) image to cover the whole space, and set the AR image at the center viewport of the BG with adjusting the transparency value of it (the AR image) to $\alpha$ in Unity.
Before the formal experiment, a simple training session is conducted for subjects to make them familiarize with the HMD and scenarios.
During the formal experiment, for each subject, 450 scenarios are randomly divided into 3 sessions with 150 scenarios per session.
Unlike previous omnidirectional saliency prediction task in VR, our work focuses more on the saliency within the perceptual viewport of AR.
Therefore, in our study, to make the perceptual viewport relatively fixed, subjects are seated in a fixed chair facing the center viewport of the BG rather than a swivel chair which is usually used for VR experiments \cite{rai2017dataset,xu2021saliency}, and they are encouraged to rotate their head freely but cannot turn their body.
The duration for viewing each superimposed scenario is set to 5 seconds.
After viewing each superimposed image, we insert a gray omnidirectional image with a red dot located at longitude = $0^{\circ}$ and latitude = $0^{\circ}$, and no AR content is displayed.
The subjects are encouraged to fixate on the red dot before the next image.
At the beginning of each session, we re-calibrate the eye-tracker to ensure the reliability of the acquired data.


\subsection{Data Processing and Analysis \label{sec:3.3}}

\hspace{0.91em} \textit{\textbf{Data Processing.}} The raw eye movement data are recorded in the format of [pitch, yaw, roll], thus we first convert the raw data to latitude and longitude coordinates.
Then we process the gaze data to extract the fixation points.
Fixation occurs when user's eyes fixate at a specific region for a short period of time.
We derive fixation by removing saccade (fast eye movement change) from the data.
Specifically, we first calculate the distance and velocity between consecutive gaze points.
Then the mean absolute deviation (MAD) \cite{voloh2019mad} in gaze position is calculated within a seven-sample sliding window (~80 ms) and potential fixations are defined as windows with a MAD less than $50^{\circ}/s$ \cite{peterson2016individual,haskins2020active}.
Fixations with durations shorter than 100 ms are excluded \cite{peterson2016individual,wass2013parsing}.
Finally, a 2D Gaussian kernel with $3.34^{\circ}$ of visual angle \cite{rai2017dataset,zhang2018saliency} is imposed to all fixations to generate the saliency map for an image.

\textit{\textbf{Qualitative Data Analysis.}}
Figure \ref{fig:3} demonstrates some sampled saliency maps of the corresponding perceptual viewport images.
Similar to the general consensus as mentioned in the introduction, we first find that a higher opacity value (\ie, higher $\alpha$, lower transparency) generally leads more attention on augmented contents, while a lower opacity value makes the background scenes more salient.
Furthermore and more importantly, we also find that visual attention in AR environment (\ie, superimposed stimuli) is jointly and significantly influenced by the AR image, BG image and mixing value.
As shown in the examples of the first two rows in Figure \ref{fig:3}, for $\alpha=0.25$, subjects tend to fixate more on salient regions of the background images, for $\alpha=0.75$, subjects tend to focus more on salient regions of the augmented images, for $\alpha=0.5$, subjects tend to fixate on the salient regions of both AR and BG images.
Thus we can clearly observe the saliency transform procedure from these three examples.
For the three examples of the last two rows in Figure \ref{fig:3}, saliency maps for $\alpha=0.25$ and $\alpha=0.5$ are more similar compared to $\alpha=0.75$, which means that the transition is slight when $\alpha$ is less than 0.5.
Therefore, Figure \ref{fig:3} qualitative illustrates that visual attention in AR environment is jointly influenced by the AR image, BG image and mixing value rather than only influenced by the mixing value.

\FigIV

\textit{\textbf{Quantitative Data Analysis.}}
As shown in Figure \ref{fig:4}, we first analyze the consistency of eye fixation distributions across subjects when the number of subjects increases.
The consistency of eye fixations between two groups is measured by calculating the linear correlation coefficient (CC).
Figure \ref{fig:4} (a) shows the CC values between sub-group (with fewer subjects) saliency maps and overall (all subjects) saliency maps, which are calculated and averaged among all stimuli.
It can be observed that the CC value, \ie, consistency, increases and converges along with the increased subject number.
Moreover, according to the consistency value and standard deviation value, we recommend that at least 10 subjects are required to obtain good consistency for conducting AR saliency experiment.
We also analyze the consistency of eye fixation distributions across subjects for stimuli in sub-categories and show the results in Figure \ref{fig:4} (b).
It can be observed that there is no obvious difference between different types of stimuli.

We further analyze the correlation between the saliency maps of the stimuli with different mixing values.
Two correlation metrics including the linear correlation coefficient (CC) and the similarity measurement (SIM) are used for measuring the correlation.
Figure \ref{fig:5} (a) shows the CC comparisons between different mixing levels for all stimuli.
The three CC values are calculated between mixing value 1 (m1: $\alpha=0.25$) and mixing value 3 (m3: $\alpha=0.75$), mixing value 1 (m1: $\alpha=0.25$) and mixing value 2 (m2: $\alpha=0.5$), mixing value 2 (m2: $\alpha=0.5$) and mixing value 3 (m3: $\alpha=0.75$), respectively, and then averaged among all stimuli.
It can be observed that the averaged CC of ``m1 \& m3'' is significantly less than that of ``m1 \& m2'' and ``m2 \& m3'', which quantitatively illustrates that the mixing value significantly influences the visual attention in AR, and saliency maps of superimposed images with middle mixing values ($\alpha$ near 0.5) tend to fuse the saliency maps of superimposed images with side mixing values ($\alpha$ away from 0.5).
Figure \ref{fig:5} (b) shows the CC comparisons between different mixing levels for the stimuli in sub-categories.
It can be observed that the averaged CC of graphic images is significantly larger than that of natural and webpage stimuli.
The reason may be that the stimuli with graphic augmented contents in our dataset have less superimposed areas compared to natural and webpage contents thus have less influence on visual attention for various mixing values.

\FigV

\section{Vector Quantized Saliency (VQSal) Prediction in AR}

As discussed above in Sec. \ref{sec:3.3}, the AR image, BG image and mixing value in a perceptual scenario jointly influence the visual attention in AR.
Thus, it is important to consider how to integrate these three types of information for saliency prediction in AR.
In this section, we first propose a vector quantization (VQ) based method for visual saliency prediction as described in Sec. \ref{sec:4.1} and Sec. \ref{sec:4.2}.
Most recent top-down saliency models, \eg, SALICON \cite{huang2015salicon}, ML-Net \cite{cornia2016deep}, SAM-VGG/ResNet \cite{cornia2018predicting}, SalGAN \cite{pan2017salgan}, rely on well-pretrained encoders to work well.
These encoders are usually pretrained on the ImageNet classification task \cite{deng2009imagenet}.
Different from these methods, in this work, we find that using unsupervised discrete representation learning as the pretraining method can lead even better saliency prediction performance.
Then, a specifically designed multi-decoder fusion network for AR saliency prediction is proposed in Sec. \ref{sec:4.3}.

\FigVI

\subsection{Learning a Discrete Representation Model with Perceptually Rich Information\label{sec:4.1}}

The procedure of the vector quantized pretraining is demonstrated in Figure \ref{fig:6} (a).
Instead of building on individuals pixels, neural discrete representation learning \cite{van2017neural} aims to represent any image $x \in \RR^{H \times W \times 3}$ by a spatial collection of codebook entries $\quantizedcode \in \RR^{h \times w \times \codebookdim}$ from the codebook $\codebook$, where $\codebookdim$ is the dimensionality of codes and $\codebook = \{z_k\}_{k=1}^K \subset \RR^\codebookdim$ is the learned perceptually rich code book.
Specifically, a given image $x$ is first encoded by the encoder $\encoder$ to get the feature vector $ \hat{z} = \encoder(x) \in \RR^{h\times w \times \codebookdim}$.
Then each spatial code $\hat{z}_{ij} \in \RR^\codebookdim$ in $\hat{z}$ is quantized by $\quantize(\cdot)$ to its closest codebook entry $z_k$ in the codebook $\codebook$ via $\quantizedcode = \quantize(\hat{z}) \coloneqq \left(\argmin_{z_k \in \codebook} \Vert \hat{z}_{ij} - z_k \Vert\right) \in \RR^{h\times w \times \codebookdim}.$
Finally, the image can be reconstructed from these codebook entries by $\hat{x}=G(\quantizedcode)$, where $\hat{x}$ is the output of the whole model.
The overall discrete representation learning pipeline is:
\begin{equation}
  \hat{x} = \decoder(\quantizedcode ) = \decoder\left(
    \quantize(\encoder(x)) \right).
\label{eq:vqrec}
\end{equation}
Since both of the codebook ($\codebook$) and the model (\ie, $\encoder$ and $\decoder$) are required to be learned.
The vector quantized loss function can be represented as:
\begin{align}
\mathcal{L}_{\text{rec}} = \Vert x - \hat{x} \Vert^2 
  &+ \mathcal{L}_{\text{perceptual}},
\label{eq:origvqloss}
\end{align}
\begin{align}
\mathcal{L}_{\text{VQ}}(\encoder, \decoder, \codebook) = \mathcal{L}_{\text{rec}} 
  &+ \phantom{\beta} \Vert \text{sg}[\encoder(x)] - \quantizedcode \Vert_2^2 \nonumber \\ &+ \beta \Vert \text{sg}[\quantizedcode] - \encoder(x) \Vert_2^2,
\label{eq:origvqloss}
\end{align}
where $\mathcal{L}_{\text{perceptual}}$ is the well-known perceptual loss \cite{johnson2016perceptual,zhang2018unreasonable}, $\text{sg}[\cdot]$ indicates the stop-gradient, $\Vert \text{sg}[\quantizedcode] - \encoder(x) \Vert_2^2$ is the
``commitment loss'' with weighting factor $\beta$ \cite{van2017neural}.

To get good reconstruction quality for this discrete representation learning, we follow the VQGAN \cite{esser2021taming} to learn a perceptually rich codebook via GAN as follows:
\begin{align}
\mathcal{L}_{\text{GAN}}(\{\encoder, \decoder, \codebook \}, \discriminator) =
  \left[ \log \discriminator(x) + \log (1 - \discriminator(\hat{x}))\right].
\label{eq:ourvqloss}
\end{align}

\FigVII

\subsection{Transfer Learning for Saliency Prediction\label{sec:4.2}}
Through the discrete representation model learned above, we can represent any image $x$ using a spatial collection of codebook entries $\quantizedcode$, and directly reconstruct the image using these codes (visual tokens) via the decoder $\decoder$.
The representation code $\quantizedcode$ includes the extremely compressed but perceptually rich information of an image, which can be directly used to decode and predict visual saliency information.
Moreover, since the decoder learned during the unsupervised discrete reconstruction process can well recover most of image information, its knowledge can be easily transferred to predict visual saliency maps and learn the saliency relationship from these visual tokens.
Specfically, in transfer learning, we freeze the feature extraction and quantization parts, and only finetune the decoder part to perform saliency prediction as shown in Figure \ref{fig:6} (b).
Our proposed VQSal model can be represented as:
\begin{equation}
  \hat{x}_{\text{sal}} = \decodersal(\quantizedcode ) = \decodersal\left(
    \quantize(\encoder(x)) \right),
\label{eq:vqrec}
\end{equation}
where $\decodersal$ is the decoder for saliency density prediction, and $\encoder, \quantize$ are frozen encoder and quantization networks, respectively.
The loss function of the saliency prediction in our paper is defined as:
\begin{equation}
  \mathcal{L} = \mathcal{L}_{\text{rec}} + \lambda\mathcal{L}_{\text{sal}},
\label{eq:vqrec}
\end{equation}
where $\mathcal{L}_{\text{sal}}=\mathcal{L}_{\text{CC}}+\mathcal{L}_{\text{KL}}$, CC and KL are two widely used metrics for measuring the accuracy of the predicted saliency maps \cite{che2019gaze}. The weighting factor $\lambda$ is empirically set as 0.2 in this paper.

\TabI

\TabII

\subsection{VQSal for AR Saliency Prediction \label{sec:4.3}}

In AR saliency prediction task, three types of images including the AR image, BG image, and superimposed image can be obtained or calculated from a scenario (see \textbf{Sec. \ref{sec:3.1} \& \ref{sec:3.2}}), and all these three components significantly influence visual attention in AR (see \textbf{Sec. \ref{sec:3.3}}, the superimposed image contains mixing value information).
Therefore, a multi-decoder fusion network based on the VQSal model is further proposed to integrate AR image information, BG image information, and superimposed image information for AR saliency prediction.
Figure \ref{fig:7} shows the overview of this VQSal-AR model.
Specifically, three images are first fed into the feature extraction and quantization modules to get visual tokens for them, respectively:
\begin{equation}
    \left\{ z_\text{AR}, z_\text{BG}, z_\text{S} \right\} = \left\{ \quantize(\encoder(x_\text{AR})), \quantize(\encoder(x_\text{BG})), \quantize(\encoder(x_\text{S})) \right\},
\end{equation}
where $x_\text{AR}, x_\text{BG}, x_\text{S}$ are input images, and $z_\text{AR}, z_\text{BG}, z_\text{S}$ are obtained visual tokens.
These visual tokens are then fed into multi-decoders and finally integrated to produce the salieny map as follows:
\begin{equation}
    \hat{x}_{\text{sal}} = \mathcal{F}(G_\text{AR}(z_\text{AR}),G_\text{BG}(z_\text{BG}),G_\text{S}(z_\text{S})),
\end{equation}
where $G_\text{AR}, G_\text{BG}, G_\text{S}$ are three decoders, $\mathcal{F}$ is the final convolution integration, $\hat{x}_{\text{sal}}$ is the predicted AR saliency map.

\FigVIII

\section{Experiments}

\subsection{Benchmark Methodology}
Although AR images, BG images, and mixing values jointly and significantly influence the visual attention in AR (see Sec. \ref{sec:3.3}), whether saliency models should consider all these parameters as input and how to calculate AR saliency accordingly still need to be discussed.
Given an AR image $I_\text{AR}$, a BG image $I_\text{BG}$, and a mixing value $\alpha$, the superimposed perceptual viewport image $I_\text{S}$ can be calculated via Eq. (\ref{eq:1}).
To get the AR saliency map $\hat{s}$, three benchmark methods for a given saliency model $\mathcal{S}$ are defined as:

\noindent\textbf{(i) Type I: only using $I_\text{S}$,} which is formulated as:
\begin{equation}
    \hat{s} = \mathcal{S}(I_\text{S}).
\end{equation}

\noindent\textbf{(ii) Type II: using $I_\text{AR}$, $I_\text{BG}$, and $\alpha$,} which is formulated as:
\begin{equation}
    \hat{s} = \alpha \mathcal{S}(I_\text{AR}) + (1-\alpha)\mathcal{S}(I_\text{BG}).
\end{equation}

\noindent\textbf{(iii) Type III: using $I_\text{AR}$, $I_\text{BG}$, and $I_\text{S}$,} which is formulated as:
\begin{equation}
    \hat{s} = \text{SVR}(\mathcal{S}(I_\text{AR}),\mathcal{S}(I_\text{BG}),\mathcal{S}(I_\text{S})).
\end{equation}

\noindent For classical saliency models, they are directly calculated on the corresponding images.
For DNN models, they are retrained on SALICON \cite{jiang2015salicon} first.
Then for $I_\text{AR}$ and $I_\text{BG}$, these DNN models are directly calculated on these images using pretrained weights to get $\mathcal{S}(I_\text{AR})$ and $\mathcal{S}(I_\text{BG})$, and for $I_\text{S}$, they are finetuned and calculated on our dataset using the protocol in Sec. \ref{sec:5.2} to get $\mathcal{S}(I_\text{S})$.

\subsection{Experimental Results \& Analysis \label{sec:5.2}}

\hspace{0.91em} \textit{\textbf{Experiments on SALICON \cite{jiang2015salicon}}.}
We first conduct experiments on SALICON to validate the effectiveness of our VQSal model.
SALICON \cite{jiang2015salicon} is currently the largest saliency dataset with 10000, 5000, 5000 images for training, validation, and test, respectively.
For fair comparison, six state-of-the-art saliency models including SALICON \cite{huang2015salicon}, ML-Net \cite{cornia2016deep}, SAM-VGG \cite{cornia2018predicting}, SAM-ResNet \cite{cornia2018predicting}, SalGAN \cite{pan2017salgan}, GazeGAN \cite{che2019gaze} are retrained on SALICON training set, and tested on SALICON validation set.
Six widely used metrics including AUC, CC, IG, KL, NSS, SIM \cite{bylinskii2018different} are used to compare the performance of these six models with our proposed VQSal model.
Table \ref{tab:1_salicon} demonstrates that our VQSal model achieves state-of-the-art performance compared to other models among all six metrics.

\textit{\textbf{Experiments on our SARD.}}
We further conduct experiments on our SARD to validate the effectiveness and superiority of our VQSal and VQSal-AR models on the AR saliency prediction task.
The benchmark study is first conducted among 13 classical saliency models including IT \cite{itti1998model}, AIM \cite{bruce2005saliency}, GBVS \cite{harel2006graph}, SR  \cite{hou2007saliency}, SUN \cite{zhang2008sun}, PFT \cite{guo2008spatio}, SMVJ \cite{cerf2007predicting}, Judd \cite{judd2009learning}, SWD \cite{duan2011visual}, Murray \cite{murray2011saliency}, CA \cite{goferman2011context}, CovSal \cite{erdem2013visual}, HFT \cite{li2012visual}, and 6 DNN saliency models including SALICON \cite{huang2015salicon}, ML-Net \cite{cornia2016deep}, SAM-VGG \cite{cornia2018predicting}, SalGAN \cite{pan2017salgan}, GazeGAN \cite{che2019gaze}, as well as our VQSal model.
Five widely used metrics including AUC, sAUC, CC, NSS, SIM \cite{bylinskii2018different} are used to compare the performance of these baseline models with our proposed VQSal-AR model.
For learning-based methods (\ie, SVR and DNN models), we divide the SARD into 5 splits with an equal number of three stimulus categories in each split and without scenario repeat.
Then we run a 5-folds cross validation experiment with 4 splits for training and 1 split for testing in each validation fold.
This 5-folds experiment can cover the whole dataset and get the prediction results for all images.
Moreover, all DNN models are pretrained on SALICON \cite{jiang2015salicon} first and then finetuned on our SARD.

Table \ref{tab:2_ar} demonstrates the quantitative results of three types of benchmark methods and our VQSal-AR model.
For classical models, Type II and Type III models generally perform better than Type I models.
Comparing the Type II and Type III for classical models, for certain models, Type III method performs better, while for other models, Type II method is more efficient.
For DNN models, Type I and Type III models generally perform better than Type II models.
The reason may be that the Type I method is finetuned on our dataset and the Type III method includes the Type I saliency map as one input feature.
Comparing the Type I and Type III for deep models, we find that for most saliency models, Type III performs better in terms of AUC and sAUC, while for other three evaluation metrics, both methods have their advantages.
Moreover, our VQSal models achieves state-of-the-art performance for all three types of benchmark methods, and our VQSal-AR model performs much better compared to all other methods in terms of almost all metrics.
Figure \ref{fig:8} also demonstrates the superiority of our VQSal-AR model.

\subsection{Ablation Analysis}
\hspace{0.91em} \textit{\textbf{Ablation for VQSal}.}
We first conduct an ablation study for the VQSal model and demonstrate the results as follows.
\begin{center}
\vspace{-.2em}
\small
\tablestyle{5pt}{1.1}
\begin{tabular}{l|cccccc}
Model $\backslash$ Metric	&	AUC $\uparrow$	&	CC $\uparrow$	&	 IG $\uparrow$	&	KL $\downarrow$	&	 NSS $\uparrow$	&	SIM $\uparrow$	\\
\shline
w/o pretraining & 0.792	&	0.606	&	34.16	&	6.147	&	1.155	&	0.594	\\
w/o $\mathcal{L}_{\text{rec}}$ & 0.837	&	0.744	&	34.46	&	5.935	&	1.432	&	0.699	\\
w/o $\mathcal{L}_{\text{sal}}$ & 0.854	&	0.854	&	34.28	&	6.067	&	1.839	&	0.744	\\
VQSal (all combined)	&	\textbf{0.863}	&	\textbf{0.869}	&	\textbf{35.18}	&	\textbf{5.434}	&	\textbf{1.863}	&	\textbf{0.766}	\\
\end{tabular}
\vspace{-.2em}
\end{center}
We first observe the significant performance drop without the vector quantized pretraining on ImageNet, which manifests that learning a perceptually rich codebook is important for transfer learning on saliency prediction to work.
Moreover, we also see that both of the $\mathcal{L}_{\text{rec}}$ and $\mathcal{L}_{\text{sal}}$ losses have significant improvement for our VQSal.

\textit{\textbf{Ablation for VQSal-AR}.}
First of all, comparing the results of the Type I method for VQSal and the VQSal-AR as shown in Table \ref{tab:2_ar}, we observe that our VQSal-AR strategy can significantly improve the performance on the AR saliency prediction task.
We further show the effect of the SALICON dataset pretraining on our VQSal-AR model as follows (note that the pretrained weights of all three decoders in VQSal-AR are obtained from the pretrained decoder in VQSal, which is pretrained on SALICON).
\begin{center}
\vspace{-.2em}
\small
\tablestyle{5pt}{1.1}
\begin{tabular}{l|ccccc}
Model $\backslash$ Metric	&	AUC $\uparrow$	&	sAUC $\uparrow$	&	CC $\uparrow$	&	 NSS $\uparrow$	&	SIM $\uparrow$	\\
\shline
w/o pretraining on SALICON & 0.895 & 0.583 & 0.853 & 2.587 & 0.719 \\
VQSal-AR & \textbf{0.903}	&	\textbf{0.591}	&	\textbf{0.893}	&	\textbf{2.687}	&	\textbf{0.758} \\
\end{tabular}
\vspace{-.2em}
\end{center}
We see a significant performance drop for our VQSal-AR model if without pretraining on SALICON, thus as mentioned in Sec. \ref{sec:5.2}, all DNN models are pretrained on SALICON and then finetuned on our dataset to get the results in Table \ref{tab:2_ar} for fair comparison.

\section{Conclusion}
Visual attention analysis and prediction are important tasks for multimedia systems.
In this paper, we mainly study human visual attention behavior in AR and its related saliency prediction task.
We first construct a saliency in AR dataset (SARD), which contains 1350 superimposed images covering 450 AR/BG scenario pairs, and a large-scale eye-tracking experiment among 60 subjects is also conducted.
Through qualitative and quantitative analysis, we conclude that visual attention in AR environment is jointly and significantly influenced by the AR contents, BG scenes and mixing values.
For better predicting saliency in AR, we propose a general saliency prediction model \textit{VQSal} and generalize it to the model \textit{VQSal-AR} for AR application.
Three benchmark methods are proposed and evaluated on our SARD, and our proposed VQSal-AR achieves state-of-the-art performance compared to these methods.

Our work considers the basic image saliency prediction task while real application scenarios are closer to the video saliency detection task due to head movements.
Our future works will extend this study and focus on predicting augmented video saliency.


\bibliographystyle{ACM-Reference-Format}
\bibliography{references}

\end{document}